# Classifying spam emails using agglomerative hierarchical clustering and a topic-based approach

Francisco Jáñez-Martino[a,b], Rocío Alaiz-Rodríguez[a,b], Víctor González-Castro[a,b], Eduardo Fidalgo[a,b] and Enrique Alegre[a,b]

[a]*Department of Electrical, Systems and Automation, University of León, Spain*
[b]*Researcher at INCIBE (Spanish National Cybersecurity Institute), León, Spain*

**ABSTRACT**

Spam emails are unsolicited, annoying and sometimes harmful messages which may contain malware, phishing or hoaxes. Unlike most studies that address the design of efficient anti-spam filters, we approach the spam email problem from a different and novel perspective. Focusing on the needs of cybersecurity units, we follow a topic-based approach for addressing the classification of spam email into multiple categories. We propose SPEMC-15K-E and SPEMC-15K-S, two novel datasets with approximately 15K emails each in English and Spanish, respectively, and we label them using agglomerative hierarchical clustering into 11 classes. We evaluate 16 pipelines, combining four text representation techniques -Term Frequency-Inverse Document Frequency (TF-IDF), Bag of Words, Word2Vec and BERT- and four classifiers: Support Vector Machine, Näive Bayes, Random Forest and Logistic Regression. Experimental results show that the highest performance is achieved with TF-IDF and LR for the English dataset, with a F1 score of 0.953 and an accuracy of 94.6%, and while for the Spanish dataset, TF-IDF with NB yields a F1 score of 0.945 and 98.5% accuracy. Regarding the processing time, TF-IDF with LR leads to the fastest classification, processing an English and Spanish spam email in 2ms and 2.2ms on average, respectively.

## 1. Introduction

Billions of spam emails are sent and received everyday[1]. All email clients have a spam folder which automatically collects undesired content that is often unnoticed. Thanks to these spam folders, unsolicited emails are less annoying and they do not swamp users' mailboxes. However, spam emails could be adapted for specific targeted attacks, jumping to our main inboxes. They may just contain advertisements and company promotions, and although this is annoying, it is harmless [8, 26].

Unfortunately, there is also a significant proportion of spam messages that have a malicious nature and whose aim could be to steal personal data, introduce malware or hijack user systems [26]. Spam generation is low-cost and identifying its creators is not a straightforward task, which makes this a very common method used by cybercriminals. In addition, the volume of spam represents a huge proportion of the total emails sent daily. According to the reports of Cisco Talos[2] and Kaspersky Lab[3], spam emails represent between 55% and 85% of the daily total volume of worldwide emails. Spam ma«y cause productivity loss, distrust in email service, annoyance or services bottlenecks which limit memory space and speed of computers, resulting in an economic expense for organisations that is steadily increasing. As a consequence of all the above, a decade ago spam was estimated to cost companies twenty billion dollars annually. This cost is likely to surpass 250 billion dollars in a couple of years [43]. Moreover, if the aim of the spam is fraudulent, the integrity, security and privacy of the user may also be exposed to cybercriminals.

Spam email is a problem widely studied in literature, since spammers constantly develop new techniques in order to bypass the email client's spam filters. Existing research on using Natural Language Processing (NLP) for spam detection has focused particularly on binary classification approaches, categorising emails into two classes, legitimate or undesired email, i.e. ham or spam [30, 5, 6, 25, 21].

---

ORCID(s): 0000-0001-7665-6418 (F. Jáñez-Martino); 0000-0003-4164-5887 (R. Alaiz-Rodríguez); 0000-0001-8742-3775 (V. González-Castro); 0000-0003-1202-5232 (E. Fidalgo); 0000-0003-2081-774X (E. Alegre)

[1]https://techjury.net/blog/how-many-emails-are-sent-per-day/- Retrieved December 2021
[2]https://talosintelligence.com/reputation_center/email_rep Retrieved December 2021
[3]https://www.statista.com/statistics/420391/spam-email-traffic-share/ Retrieved December 2021

It is a well-known fact that spam can be classified into different categories[4]. In addition, some spam categories may be more harmful than others, and some of them may be more prone to go through spam filters undetected. Therefore, it would be a valuable improvement to detect not only if an email is spam, but also its type. The multi-classification of spam emails could improve the cybersecurity incidents handling, companies and citizens protection and early warning by identifying the behavioural patterns of spammers as a vital aspect of spam detection [19]. Due to the malicious nature of some of the spam emails, it is important to analyse its content to prevent cyber-attacks or campaigns against specific targets [24, 46]. At the time of writing this paper, there is only one work carried out by Murugavel and Santhi [44] that deals with multiple threads of spam from a text analytic perspective, but without applying artificial intelligence.

In this paper, we approach the spam email problem from a different and novel perspective. We analyse the text content to identify cybersecurity topic-based class detection. These classes emphasise the most common topic hoaxes that citizens and companies have to face when they receive spam daily. Since we gain insight into the spam email data, cybersecurity organisations may identify campaigns more easily in relation to the scam topic and enhance the warnings against them. The main contributions of this work can be summarized as follows:

1. We carried out an analysis and investigation of the textual part of spam emails using a hierarchical clustering algorithm in order to divide them into classes based on a cybersecurity topic-based approach.
2. We presented an email preprocessing method to extract the textual content from spam emails considering spammer tricks such as (i) introducing part (or all) of the spam message into images and (ii) hiding random text in the body of the email (known as "salting").
3. We created a novel dataset called Spam Email Multiclassfication (SPEMC) that is divided into two subdatasets: one with emails in English and another with emails in Spanish. Each one contains almost 15K spam emails labelled into a predefined set of eleven categories.
4. We introduced a framework to classify spam emails into cybersecurity categories using machine learning and natural language processing techniques. The proposed approach can be integrated into tools and services whose objective is to serve citizens and organisations, helping them to identify harmful spam, like the one containing extortion hacking, fake reward, identity fraud or false job offers.

Our collaboration with the Spanish National Cybersecurity Institute (INCIBE)[5] aims at developing solutions based on machine learning that could be useful for Public Administrations, Industry or Law Enforcement Agencies. This work is an extension of a preliminary study [32] and it has been influenced by some research carried out about the dark web ([10, 3, 9, 27]), where domains in the onion router (Tor) darknet are classified in multiple categories depending on their contents, instead of just dividing them as legal or suspicious of being illegal. The ultimate goal of this work is to enable the extraction of meaningful information from large amounts of undesired -and possibly harmful- spam emails, which can help Law Enforcement Agencies (LEAs) or companies to fight against them.

To the best of our knowledge, there are no works that tackle the spam email problem from a cybersecurity topic-based perspective using NLP and machine learning. For the first time, in this paper we propose SPEMC-15K-E and SPEMC-15K-S, two novel datasets containing approximately 15K spam emails each. The SPEMC datasets have been semi-automatically labelled into 11 categories by means of an agglomerative hierarchical clustering dealing with the hidden text problem efficiently. SPEMC-15K-E and SPEMC-15K-S comprise English and Spanish spam respectively, which are the second and third most spoken languages in the world[6]. Besides, we propose a spam multi-classification pipeline, assessing sixteen different combinations of encoding techniques with machine learning classifiers for English and Spanish spam emails, setting baseline results for the SPEMC datasets for future research on the cybersecurity topic.

In addition, in order to extract all valuable text from the spam email, we detect two well-know spammer tricks in our datasets and propose a solution to minimise their impact in the classification. Before encoding the content of the undesired email, we also introduce a different way of working with the text included in images and the salting. For the spam that includes images, we extract the text using OCR techniques, instead of ignoring it. For the hidden disturbing text, instead of looking into the HTML tags, we extract the text which is visible to the user using OCR technologies. An overview of the entire process, including the creation of the datasets, is shown in Fig. 1. In addition, a flow chart explaining the process conducted is depicted in Fig. 2.

The process of knowledge discovery in this field can be simplified with the automatic classification of spam emails into several categories. This multi-classification model can help current cybersecurity agencies that manually analyse spam emails using hard-coded rules, trying to avoid loss of work productivity, malware distribution [15] and phishing [16] as well as to detect cybercrime campaigns [24].

---

[4]https://encyclopedia.kaspersky.com/knowledge/types-of-spam/ - Retrieved December 2021
[5]https://www.incibe.es/en Retrieved December 2021
[6]https://www.ethnologue.com/, Retrieved December 2021

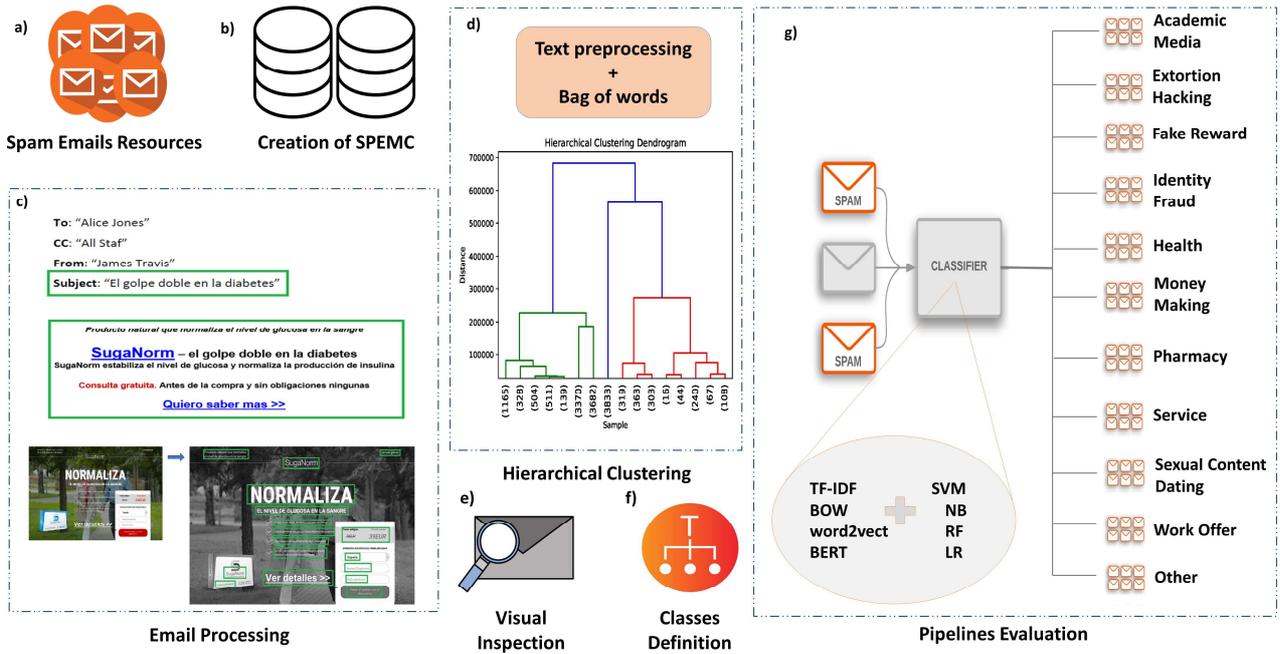

**Fig. 1:** Spam email multi-classification process: a) extraction of 15K random spam emails per language from resources, b) pre-process emails, c) extraction of all visible text of every email, d) text preprocessing on each email, then encoding with Bag of Words and finally, hierarchical clustering, e) manual review of the clusters, f) category labelling, g) training and evaluation of 16 pipelines of text classification.

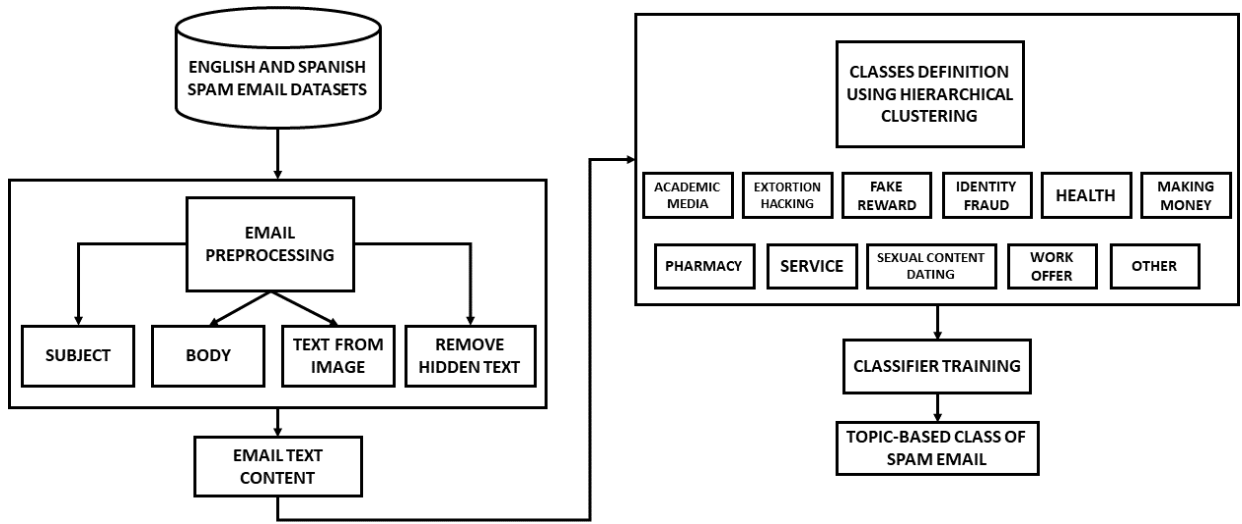

**Fig. 2:** Flow chart of the entire process to develop our proposed model capable of detecting cybersecurity topic-based classes automatically.

The rest of the paper is organised as follows: related works are reviewed in Section 2. Section 3 explains the methodology we have followed to create the SPEMC datasets. The set of the designed classification pipelines is explained in Section 4. After that, in Section 5, we detail the experimental setup and we discuss the results. Finally, Section 6 presents our conclusion and future work.

## 2. Literature Review

### 2.1. Spam email detection

Organisations and researchers have been developing filters to classify emails as spam or not spam for the last few decades. Models based on machine learning and NLP have become the state-of-the-art filters. Barushka and Hayek [6] used a deep learning model and, later, Faris et al. [25] developed a genetic algorithm as a feature selector along with a Random Weight Network. Recently, Saidini et al. [49] combined semantic features extracted by different text encodings, e.g. doc2vec or Bag of Words (BOW), with six machine learning algorithms, such as Support Vector Machine (SVM), k-nearest neighbour, Adaboost, Naïve Bayes (NB), decision trees and Random Forest (RF) to identify spam emails. Dedeturk et al. [22] created a filter using an artificial bee colony as feature selector and Logistic Regression (LR) as classifier. Despite the impact of the deep learning models in many tasks, Mekouar [39] recommended traditional algorithms like RF and NB due to their performance in spam detection.

Although the binary classifiers developed recently show high performance, it is worth highlighting that the emails used to calibrate the machine learning models come from publicly available datasets that are dated from the earliest 2000s. For example, Barushka and Hayek [6] obtained remarkable results on publicly available datasets like SpamAssassin[7] or the Enron-Spam dataset [40]. Faris et al. [25] also assessed their binary spam filter on the SpamAssassin dataset. However, it is important to highlight that spam email has a changing nature due to time (evolution of subjects) and to the techniques used by spammers wishing to elude spam filters, what inevitably leads to shifts in the dataset [48]. Due to this fact, the most recent works in spam email are training their models without considering current spammer tricks. Bhowmick and Hazarika [8] enumerated a list of the most popular spammer tricks, among them, the use of image-based spam and insertion of random text in the email body. The former consists on inserting the spam message inside an image attached to the email to bypass the filters based on textual analysis. There are some works [12, 51, 37] that have dealt with detecting spam emails through classifying the attached images. They used machine learning models and the image properties, e.g. the metadata or the colour, as features. Other works, like [45], handled the image-based trick from an Optical Character Recognition (OCR) perspective to recognise and extract the letters and words from a spam image [19].

Spammers also try to confuse the textual spam filters by inserting pieces of random text inside the email body and hiding it conveniently, e.g. by reducing the font size or by making it invisible to readers. This trick, which normally uses HTML tags, is known as hidden text or salting [35].

There are a small number of works oriented to detect the salting trick in spam emails and use this content to enhance binary spam classifiers, [7, 35]. They try to identify if a character is hidden text by analysing its visibility, i.e. checking out anomalies in terms of colour or size, presenting an introduction of OCR solution. Despite being a common problem nowadays, it is often overlooked, and we have found no more works that deal with this problem.

### 2.2. Topic-based detection in the spam field

In their study about spam opinion detection, Ligthart et al. [34] concluded that, in some scenarios where binary classification is inadequate, multi-class classification is required, and defined this task as a demanding challenge with high research efforts. To the best of our knowledge, a few works have addressed the multi-classification based on topic-based approach in spam email [44, 49].

Saidini et al. [49] divided both spam and not spam emails into six pre-defined domains according to the topics of most common advertisements, e.g. computer, adult, education, finance, health and others. They developed a model based on machine learning and natural language processing to detect these domains and perform a binary classification of an email.

Muragavel and Santhi [44] identified seven spam categories – or threads – through the count of the most frequent words in a dataset of emails. They also provided some statistics, concluding that the most frequent thread on spam

---

[7]https://spamassassin.apache.org/old/publiccorpus/ Retrieved December 2021

emails is promotional advertisements. However, they did not use either NLP or machine learning techniques to assign an email into a category. Besides this, their work in the multi-classification problem of spam is based on a small dataset, which comprises 1040 emails, 842 of which are spam, which is not large enough for providing consistent statistics, nor for building a robust pipeline for automatic spam classification. Indeed, they also pointed out the image-based spam trick, but they did not mention how to make use of this information or how to classify these emails automatically.

However, the previous works did not tackle the spam email problem from a cybersecurity topic-based perspective using NLP and machine learning.

### 2.3. Hierarchical text clustering

The hierarchical clustering methods are divided into agglomerative and divisive, bottom-up and top-down approaches [1]. An agglomerate clustering starts with single-point clusters and, according to their similarity, recursively merges two or more clusters until achieving a stop criterion. These properties allow the algorithm to find out high level relationship among categories and join related clusters with fewer number of examples. Some works have been used the hierarchical clustering in textual tasks due to the versatility and ease to filter the data visually. Al-Mahmoud et al. [4] evaluated a hierarchical algorithm in the task of text clustering and concluded that its high performance did not degrade depending on the number of documents and number of clusters. In their work, De Campos et al. [20] assessed text clustering techniques and remarked that the hierarchical algorithms work quite well for filtering problems. Mahdavi et al. [36] carried out their investigation to discover relationships among dataset entities using hierarchical clustering to analyse the datasets.

## 3. Datasets: SPEMC-15K-E and SPEMC-15K-S

At the time of writing this paper, there is only one work by Murugavel and Santhi [44], which tackles the spam email problem as a multi-classification problem. They identified seven categories or threads using the count of the most frequency words on a dataset which contains 1040 emails, 842 of which are spam. The categories are chain letters, email spoofing, promotional advertisements, hoaxes, malware warning and porn spam. They provided some statistics, number of emails per category and most and less representative class. This dataset does not count with numerous emails to provide a consistent statistics and train a robust automatic classifier.

We perform our experimental study on two novel and present-day datasets: SPEMC-15K-E (Spam Email Classification dataset - English) and SPEMC-15K-S (Spam Email Classification dataset - Spanish), containing approximately 15K spam emails each one. The purpose of this work is to provide a real solution for the INCIBE environment to enhance the security and privacy of companies and citizens. We address the study of spam emails in English and Spanish because they are the most reported languages by INCIBE. The analysis of the spam email leads us to a division into eleven spam categories.

We defined these categories using machine learning and NLP techniques and with the supervision and support of an expert group of technicians from INCIBE. We focused on the spam email topics from a user point of view, both companies and citizens. We wanted to differentiate what is the bait of the hoaxes, advertisements or chain letters emails in order to detect, with further information, campaigns of spam emails, such as miracle product scams.

### 3.1. Datasets Creation

Since both datasets were created following the same process, contain the same number of categories and almost the same number of emails, and the only difference between them is the language - **E**nglish or **S**panish - we use the abbreviation SPEMC-15K to name them. To build the SPEMC-15K datasets, INCIBE provided us the spam emails, which were previously collected by honeypots of the Spanish national research and education network called RedIRIS [8].

We had a total of 70K emails from November and 15K from April of 2019. Once we received the data, we first extracted 15K random emails per language, i.e. English and Spanish, from the initial spam collection provided by INCIBE. In general terms, we used an agglomerative hierarchical clustering to divide each dataset into clusters in order to identify a hierarchy of topic-based groups inside. We extracted 15K emails per language due to the limitations of hierarchical clustering for large datasets, such as high computational time and space complexity [36], as well as human resources to analyse the cluster outputs. The use of 15K emails allowed us to find a trade-off between time and space complexity and an appropriate experimentation. Later, an expert carried out the annotation of each group and checked that the division was suitable, while cybersecurity experts from INCIBE supervised the annotation and definition of the classes. All authors of the paper validated the previous annotations during the entire process.

---

[8] https://www.rediris.es/index.php.en Retrieved December 2021

In more detail, we applied the email processing (Section 4.1) and the text preprocessing (Section 4.2) steps and we encoded their text using a BOW model [33]. To obtain a first division of the unlabelled email corpus, we followed Biswas et al. [10] work for building a Tor (The Onion Router) image dataset and Zhang et al. [53] work for dividing spam images into clusters using agglomerative clustering. We clustered the BOW feature vectors through an agglomerative hierarchical clustering, evaluating different linkage metrics. The Ward's minimum variance [31] appeared to be the most suitable linkage approach due to getting a larger and faster separation between clusters.

Finally, we manually selected a cut-off distance by observing the resultant dendrograms. We established an approximate range of possible categories based on experts' suggestions from INCIBE and their preferences. We obtained 16 clusters for English and Spanish. We visually inspected all the emails from every cluster to assign an initial tag that helped to define the category later on. After merging some similar clusters into the same class and looking for the same categories for both languages, we obtained the final categorisation with 11 classes labelled. INCIBE experts checked out our labelling in order to advise us and define a suitable list to contemplate the interests of companies and citizens. We sought to automate a categorization task that is carried out by experts manually, to overcome time and resources limitations. Due to this fact, the experts' help allowed us to determine classes according to their needs, using the clusters as a baseline. Although we can relate every cluster with a class after some reassignments of emails with the help of experts, a visual inspection is recommended to be sure about the type of emails found in every cluster. Unfortunately, we cannot make both datasets publicly available since they do not belong to us (i.e., they were provided by INCIBE), and they contain personal information, which is difficult to anonymise entirely in a reliable way. Fig. 3 shows the dendrograms for both languages. We can observe a close distance between similar topics and writing styles, such as Academic Media, Health and Pharmacy, Service and Work Offer. Likewise, between Extortion Hacking and Identity Fraud. Although the majority class - Sexual Content Dating - groups several clusters, the content of each one follows the same topic and purpose, and a subdivision could have added noise to the model.

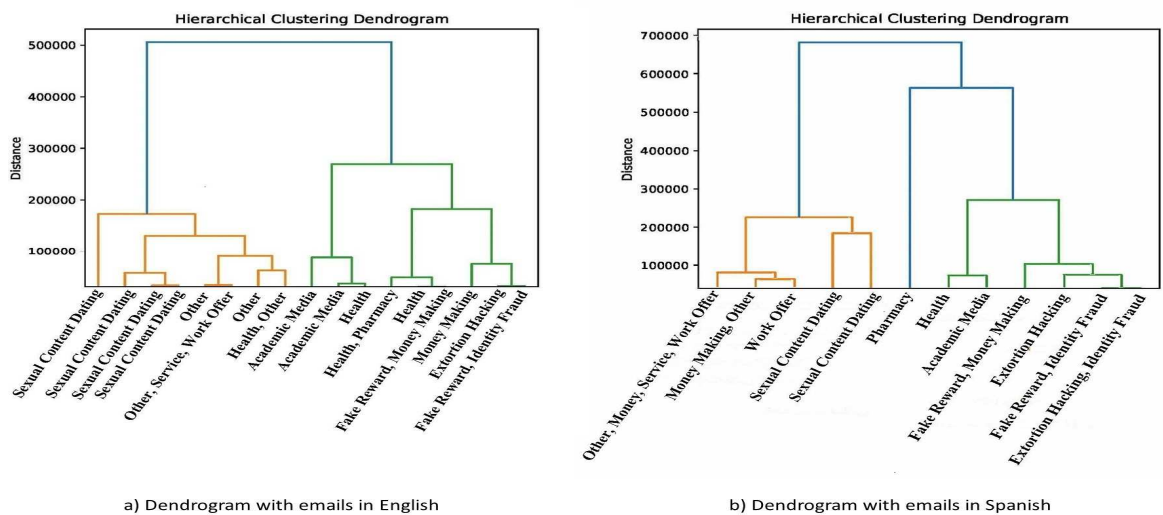

**Fig. 3:** Dendrograms provided by hierarchical clustering of both languages English and Spanish. Axis X represents the class associated to every cluster, and Axis Y represents the distance among the clusters.

### 3.2. Datasets Characteristics

The SPEMC-15K datasets contain the following classes: *Academic Media, Extortion Hacking, Fake Reward, Health, Identity Fraud, Money Making, Pharmacy, Service, Sexual Content Dating, Work Offer* and *Other*. Next, we briefly describe each one of the spam classes, Table 1 shows fragments of email messages representative of each spam category and Figures 4 and 5 depict a word cloud per category in SPEMC-15K-E and SPEMC-15K-S, respectively. Table 2 shows the number of emails and the proportion of each class for both datasets.

**Academic Media** includes spam emails related to scientific conferences or journals and education services such as masters, seminars or courses. The English emails which belong to this class are mainly focused on the scientific community, and its language is rigorous and formal, emulating real conferences and calls for papers. Apart from including these emails, the Academic Media Spanish emails mostly involve courses for personal skill development.

**Extortion Hacking** contains emails which request a payment from the users in exchange for the sender does not reveal private content of them. The language used is formal without orthographic and grammatical errors in order to scare the victim being as real and severe as possible. Generally, the emails follow a similar structure and use similar words, varying the threat.

**Fake Rewards** covers emails where the sender offers an unexpected recompense for the receiver; a famous example is the Nigerian Prince scam. In general, these emails depend on the reward, and the more valuable is the reward, the longer and more explanatory is the message.

**Health** is related to miracle pieces of advice, products and news which improve the user well-being. These emails attempt to convince the user through close communication and emphasising the importance of what the message promotes. They address a large variety of topics related to health problems, such as sexual, physic or psychological.

**Identity Fraud** includes emails whose sender attempts to pose as a well-know company by using its name and brand, or a person who sends an email very similar to ham email. They sometimes trick users by looking like an email with a wrong receiver in order to achieve a naive response from the victim. They also use social engineering techniques, building phishing emails to obtain private information from the victim. Due to these characteristics, cybersecurity experts point out identity fraud as one of the most harmful classes for companies and citizens.

**Money Making** is composed of emails which offer online services to earn money quickly, such as casinos, fast tricks to earn easy money or betting shops. Money Making emails often use a careless and repeating structure, looking for highlighting the ease of gaining money.

**Pharmacy** includes the sale of many known drugs via the Internet. They emulate a real pharmacy by listing their products and trying to create an email which transmits confidence.

**Service** covers the emails with advertisements of Small and Medium-sized Enterprises (SMEs) or personal services. They promote a profession to solve specific user problems.

**Sexual Content Dating** contains sex web pages and sex propositions. A dating email often uses the same structure, just changing some words, and show explicit sexual content. On the other hand, the propositions are generally a more careful and smart language in order to convince the user with something else than just sexual arguments.

**Work Offer** groups spam emails which offer a fake job with significant benefits to the user. These emails usually follow a pattern and only differ in the work conditions.

**Other** mainly contains disclosures, discoveries and products about politics, economy and technology. These emails are similar to Health class emails, and the major difference between them is the topic.

## 4. Methodology

We divided the automatic classification of spam emails in two stages: *email processing*, where we extracted the textual information from raw emails, and *text classification*, where we pre-process, encode and classify the textual information.

### 4.1. Email processing

One of the main purposes of our research is to classify spam emails based on its topic. Generally, emails are divided in two parts: (i) header and (ii) body, which comprises text, multimedia objects and attachments. Since we are working in the NLP field, we focused on those elements of each email where we can extract textual information.

From the header, we extract only the field *Subject*, which usually summarises the content of the body into a few words, and we do not use the rest of the fields of the header, such as CC/BCC or address email. Although they are suitable to detect spam emails, since our objective is to classify the spam email into a default set of categories related to its topic, we only used the subject field because it contains the greater amount of textual information compared to other headers of the email.

Traditionally, the *email body* had plain text without formatting options. Nowadays, emails usually are coded in HTML format, which allows enhancing the email design through the use of templates, images and extra functionalities [14]. However, some emails contain both formats – plain and HTML – to ensure that the client can read the email without depending on the service.

**Table 1**
Piece of an email for every spam class defined.

| Class | Example |
|---|---|
| Academic Media | Better Packaging Better Living. Join FSQ Europe 2020 taking place on 29th 30th January 2020 in London, UK and hear senior representatives of British Plastic Federation, OFI Technologie Innovation GmbH and Client Earth give presentations in Session 2 in the morning of Conference Day 1 entitled "Better Packaging,Better Living" focusing on |
| Extortion Hacking | I've been watching you for a few months now. The fact is that you were infected with malware through an adult site that you visited...I made a video showing how you satisfy yourself in the left half of the screen, and in the right half you see the video that you watched. With one click of the mouse, I can send this video to all your emails and contacts on social networks. I can also post access to all your e-mail correspondence and messengers that you use. If you want to prevent this, transfer the amount of $732 to my bitcoin address. (if you do not know how to do this, write to Google: "Buy Bitcoin"). |
| Fake Reward | I am interested to transfer and invest in your country through your assistance.I am in Ghana presently and I have the sum of Ten Million Eight hundred thousand US Dollars which I would like to transfer into your account and invest in your country if possible. |
| Health | 18 months ago, I discovered a weird method that can safely and naturall improve your hearing , no matter how complicated your hearing problems are. So far, it's already helped over 96, 623 people who found this brilliant, ear-saving method to save them from going DEAF... |
| Identity Fraud | Dear, Please find attached copies of documents that we were sent the original ones. Thanks & Regards, PEF PVT LTD |
| Money Making | SlotoCash Casino Trusted Online Since 2007 ExclusiveOffer Get $31FREE NoDepositRequired Code:31FREE 200% MatchBonus+100FreeSpins EnterPromoCode:SLOTO1MATCH |
| Pharmacy | Online Pharmacy, Guaranteed Quality! Save your money, time, efforts. You'll never find better offer! Best medications available are sold at our trusted online pharmacy! This month at half price! - Fast World shipping - Secure ordering - Lowest price - NO PRESCRIPTION REQUIERED |
| Service | Dear Sir/Madam, Nice day, Glad to hear you are in inflatable outdoor products market! We are Supplier for air track, inflatable SUP board, inflatable sport game ,inflatable water park, we also have very popular style, it is Inflatable gymnastic tumbling mat. |
| Sexual Content Dating | Want sex tonight, and new pussy every day? Here you can find any girl for sex! They all want to fuck |
| Work Offer | Hello! We are looking for employees working remotely. My name is Anderson, I am the personnel manager of a large International company. Most of the work you can do from home, that is, at a distance. Salary is $3500-$7000. If you are interested in this offer, please visit Our Site Best regards! |
| Other | AutoCharge² - Magnetic Phone Holder and Charger Special 50% Black Friday Sale - Order Now at 50% Off! Hat |

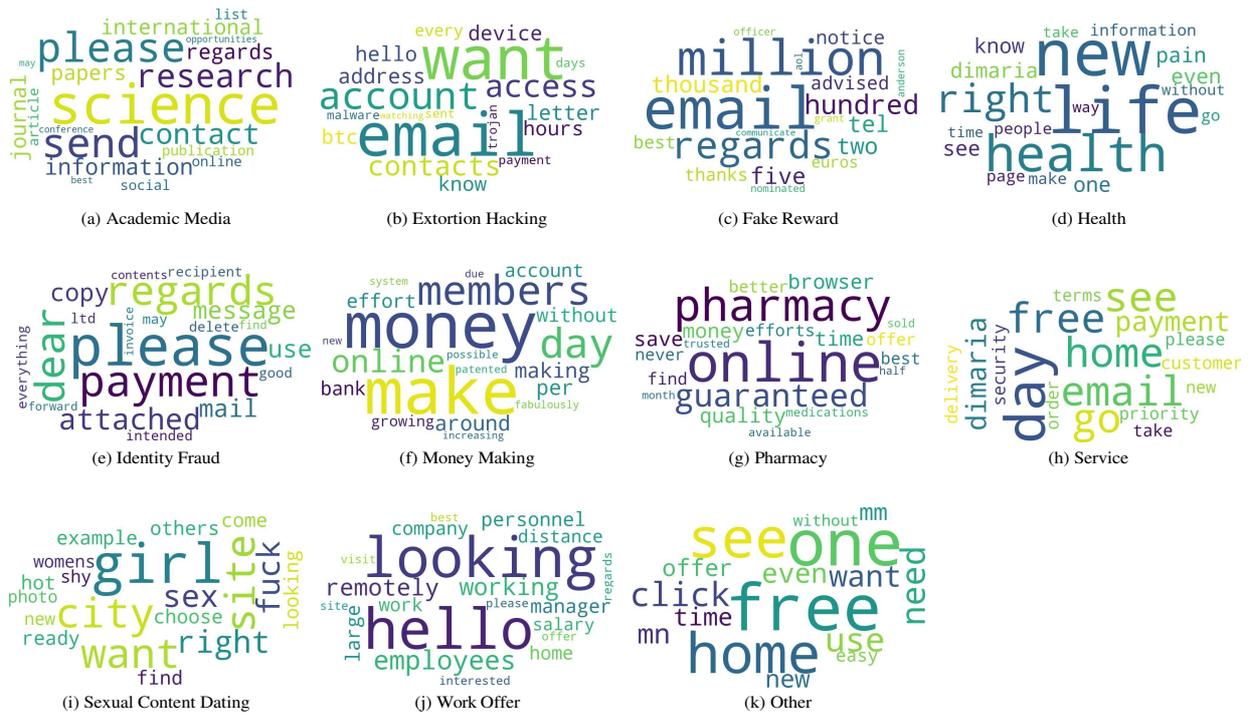

**Fig. 4:** The top 18 most frequently used words of every class in SPEMC-15K-E depicted in a word cloud.

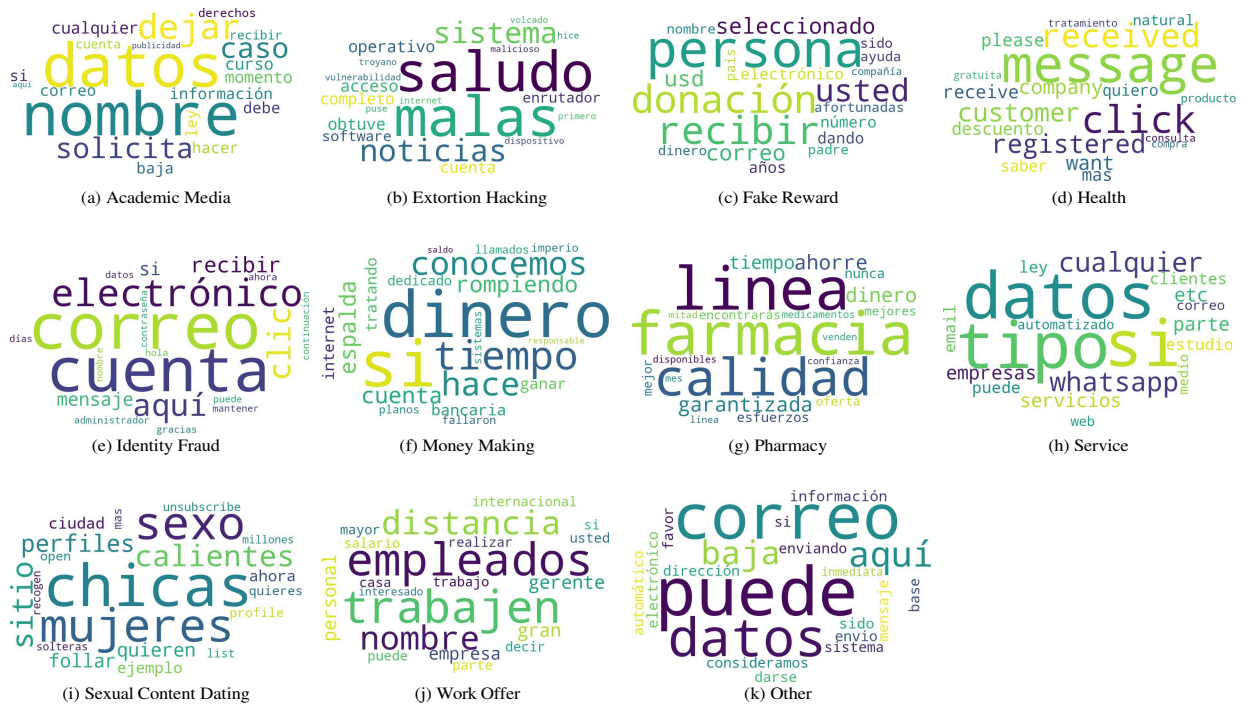

**Fig. 5:** The top 18 most frequently used words of every class in SPEMC-15k-S depicted in a word cloud.

**Table 2**
Number of emails and the percentage per class and dataset

| Class | SPEMC-15K-E | | SPEMC-15K-S | |
|---|---|---|---|---|
| | Count | % | Count | % |
| **Academic Media** | 64 | 0.44 | 690 | 4.60 |
| **Extortion Hacking** | 197 | 1.36 | 259 | 1.73 |
| **Fake Reward** | 240 | 1.66 | 37 | 0.25 |
| **Health** | 2499 | 17.25 | 490 | 3.27 |
| **Identity Fraud** | 334 | 2.31 | 144 | 0.96 |
| **Money Making** | 434 | 3.00 | 513 | 3.42 |
| **Pharmacy** | 17 | 0.12 | 3833 | 25.57 |
| **Service** | 183 | 1.26 | 271 | 1.81 |
| **Sexual Content Dating** | 7924 | 54.73 | 7062 | 47.10 |
| **Work Offer** | 55 | 0.38 | 1165 | 7.77 |
| **Other** | 2532 | 17.49 | 528 | 3.52 |
| **Total** | 14479 | | 14992 | |

To process the text from the email body, we consider three scenarios: (i) emails with only plain text, (ii) with only HTML format and (iii) both simultaneously. In emails with both formats, we prioritise the analysis of the HTML one, rather than the plain text. We do that because the HTML format characteristics also give to the spammers a more sophisticated tool to enhance their tricks [8, 48, 26].

Particularly, we found out emails with hidden text within, which would affect the performance of a text classifier due to the introduction of random text, invisible for the users, but utilised by the spam filters to detect a spam email. Although this trick might impact spam detection, to the best of our knowledge, it has not been taken into account during the last few years. For this reason, we recover the use of hidden text from the latest research available ([35],[7]). First, we convert the HTML email body into an image containing the entire email. Then, we transform the image to greyscale, and finally, we extract the visible text from this image by using an OCR. It is worth emphasising that we assume every spam email with HTML part is suspicious to contain hidden text. Following this methodology, we ensure the extraction of all the visible text seen by email clients, allowing the information to be classified in the same way a human being does.

According to Bhowmick et al. [8], spammers avoid filters based on textual content attaching images containing text, instead of writing in the email body. Researchers have detected the image-based spam by analysing the attached image [51, 12]. To consider the textual information extracted from an image, we also applied an OCR to extract the text embedded in those images.

After checking out the language of the text obtained from each part, i.e. the subject, body and images attached, we joined all the text to be processed, as a whole, in the Text Classification stage. If the text is not written in English or Spanish, it is discarded. Since INCIBE is a Spanish organisation, they are interested in both languages by being the most harmful for the Spanish companies and citizens. We only use emails where the language of all these parts is the same, to avoid that the future classifier performance would be impacted negatively. This step is vital, avoiding emails whose content is in several languages, e.g. English images with text body written in Russian.

### 4.2. Text Classification

This stage is divided into three phases, following other works like [47]: text pre-processing, representation and classification. In the pre-processing, first, we removed single characters, numbers and letters. If there are characters or numbers inside a word, we eliminated them. Then, we changed the text to lowercase, and finally, we removed the stop words, the duplicated words and tokenized the resulting text. We have not applied a stemming method due to their ambiguity, which could be a cause of a wrong classification in a misleading environment as the spam email is.

To represent the text, we selected two popular techniques based on word frequency – i.e., Bag of Words (BOW) [29] and Term Frequency - Inverse Document Frequency (TF-IDF) [2] – together with two recent word embedding techniques: word2vec [41, 42] and Bidirectional Encoder Representations from Transformers (BERT) [23].

It is worth highlighting that BOW and TF-IDF allow straightforward implementations with low computational requirements. However, they do not consider the words' order. BOW [29] represents a text corpus by means of a feature vector whose components are the frequency of each word. TF-IDF [2] builds a sparse vector assigning a numerical value to each word of the text corpus, emphasising it when a word appears many times in a text and fewer times in the rest of the corpus. On the other hand, word2vec and BERT represent a text as a vector which encodes the relationship between the words, i.e. their context. This enables similar words to be represented closer in the embedding space, enhancing the semantic analysis and context of the words. These techniques build a vector with lower dimensionality than traditional methods, being able to manage large datasets without spending many computational resources. However, the models have a larger size than the word frequency models, which might be a drawback for a real-time application.

Word2vec [41, 42] tries to maximise the likelihood that words are predicted from their context, with a Continuous Bag of Word (CBOW) model, or vice versa, skip-gram model. BERT [23] is based on the context, taking as baseline a masked language model and pre-trained using bidirectional transformers [50], and it encodes words using bidirectional instead of unidirectional representations. We selected word2vec and BERT because they are the most significant word embedding with different approaches. The model word2vec is based on learning context-independent word representations, whereas BERT relies on learning context-dependent word representations.

Finally, we combined every text representation with each of four well-know machine learning algorithms, Support Vector Machine (SVM) [17], Näive Bayes (NB) [38], Random Forest (RF) [11] and Logistic Regression (LR) [18], resulting in 16 different pipelines or trained models for the task of Text Classification.

## 5. Empirical Evaluation

### 5.1. Experimental Setting

We carried out our experiments on a personal computer with an Intel(R) Core(TM) $i7-7th Gen$ with 16G of RAM, under Ubuntu 18.04 OS and Python 3.

We assessed several multi-classification models on the two datasets presented in this paper, SPEMC-15K-E and SPEMC-15K-S (see Section 3), which contain spam emails in English and Spanish, respectively. Regarding the process of building the datasets, we detected the email language using the Python module langdetect[9] and implemented the agglomerative hierarchical clustering algorithm with the Python3 library scipy[10]. We extracted the text from images and email HTML image through a Python wrapper of tesseract-ocr[11], called pytesseract[12].

Both datasets are highly imbalanced, finding that the majority class in SPEMC-15K-E, Sexual Content Dating, contains 7924 emails whereas the minority class, Pharmacy, only has 17 emails. Similarly, the number of elements in the SPEMC-15K-S dataset in the majority and minority classes are 7062 and 37 emails, corresponding to Sexual Content Dating and Fake Reward, respectively. To address this class imbalance, we assigned a proportional weight for each class depending on its number of emails by using the class-weight parameter in scikit-learn Python library[13]. We used scikit-learn to implement the pipelines and nltk[14] to remove the English and Spanish stopwords.

For the text representation step with BOW and TF-IDF, we selected a vocabulary size of 7000 and 10000 words, respectively. Regarding the minimum number of appearances per word for the English and Spanish dataset, they were set to 5 and 3, respectively. Spanish verb conjugations were a challenge, and we had to design a preprocessing step without the stemming and lemmatization techniques. Consequently, we considered a fewer number of word appearances to create a robust and wide vocabulary in Spanish vectorizers.

We built a doc2vec encoder based on word2vec model provided by gensim[15]. The doc2vec model is the sum of all word vectors that compound the email text. We trained the doc2vec model during 10 epochs with an alpha value of 0.025 and the size of the doc2vec vector, which represents each email, is 100 elements. We selected 'distributed memory', i.e. DBOW option, to preserve the order of the words. We trained the word2vec model with a vocabulary of 15K words per language, i.e. English and Spanish, extracted from the emails in order to import the relation between words in a spam context due to its difficulty for handling words that have never seen before. The rest of the word2vec and doc2vec parameters were set to default values. We implemented BERT by means of a client-service[16]. After an empirical evaluation, we chose the best configuration of pre-trained BERT models. Thus, for English pipelines, we chose a BERT model with 24 layers, 1023 hidden layers, 16 heads and 340M parameters, only trained with English vocabulary and for Spanish pipelines, we selected a multi-language BERT model, which was pre-trained in 104 languages, with 12 layers, 768 hidden layers, 12 heads and 110M parameters.

---

[9]https://pypi.python.org/pypi/langdetect Retrieved December 2021
[10]https://www.scipy.org/ Retrieved December 2021
[11]https://github.com/tesseract-ocr/tesseract Retrieved December 2021
[12]https://pypi.python.org/pypi/pytesseract Retrieved December 2021
[13]https://scikit-learn.org Retrieved December 2021
[14]https://www.nltk.org/ Retrieved December 2021
[15]https://radimrehurek.com/gensim/ Retrieved December 2021
[16]https://github.com/hanxiao/bert-as-service Retrieved December 2021

For the classification step, we show below the parameter tuning per model, and the rest of the model parameters were left with their default values. We took a "One Vs Rest" (OVR) approach for all the classifiers. We selected a linear kernel for the SVM model, tuning the $C$ value. The $C$ parameter is an optimiser for both classifiers: a high value looks for a lower margin of hyperplane separation. Regarding NB, we used a Multinomial distribution for the frequency-based encoders, i.e., TF-IDF and BOW. Due to the incompatibility between negative values of word embeddings, i.e., word2vec and BERT, and the Multinomial distribution, we set a Gaussian distribution for these cases. For the RF model, we set the number of trees to 250. Lastly, we chose a $C$ value of 1000 and 120 as the maximum number of iterations for the LR model.

We evaluated the performance with 10-fold cross validation, reporting accuracy, precision, recall and F1 score. Despite working with imbalanced datasets, we assumed that every class has an equal actual value, and thus, we evaluated every model by means of the macro-average. We seek to classify spam emails without depending on their overall proportion in the dataset. This metric globally aggregates the contributions of each class, considering all classes with the same weight to calculate the average metric. Macro-average is considered more suitable when there are small-size classes [28, 52]. Additionally, we also obtained the micro-average, which evaluates every class individually, and the weighted-average, which considers the support of each class.

We also report the average processing time per email for the entire pipeline. Runtime is an important parameter for converting this solution into a real-world application, due to the massive number of spam emails that are processed on a daily basis.

Finally, we selected the most adequate pipeline per language by analysing jointly the F1 score, accuracy and execution time.

### 5.2. Experimental Results

Spam emails were labelled according to their topic into eleven categories. Our purpose is to provide a solution based on machine learning and NLP in order to automatically analyse spam emails, and give support to Cybersecurity Institutes. For that reason, we combined four text representations with four classifiers, resulting 16 pipelines to automatically categorise, for the first time in the literature, spam emails into several categories.

Table 3 shows the performance of every pipeline in terms of Macro precision, macro recall, macro F1 score, accuracy and runtime per email.

### 5.3. Discussion

In Table 3, it can be seen that the combination of TF-IDF and LR obtains the highest performance for English spam multi-classification, with a Macro F1 score of 0.953 and an accuracy of 94.6%. For Spanish multi-classification, the combination of TF-IDF along with NB depicts the best performance considering a Macro F1 score of 0.945 and an accuracy of 98.5%.

Regarding the runtime, the combination of TF-IDF with LR achieved the shorter execution time in both languages, classifying an English or Spanish email in an average of 2ms and 2.2ms, respectively. SVM combinations are the slowest among the evaluated pipelines, with times from 8.7ms to 98.6ms.

In a multiclass setting, micro-averaged precision and recall take the same value and it turns out to be identical to accuracy. Regarding the weighted-average, we observed they are close to the micro-average due to the fact there is a clearly majority class in both languages.

Although they are the highest performance pipelines, the accuracy of most pipelines is over 89.0%. The pipelines based on word2vec obtained the lowest results, which means the spam email dataset used to train the model is not suitable for this purpose. We trained our word2vec models as a doc2vec model from scratch using a small dataset for both English and Spanish language, which only included tens of thousands words belonging to email documents. A short vocabulary and similar context may be the main drawbacks to establish a robust relation among words, which may produce word vectors with close values and similar predictions to be assigned to the wrong class.

Moreover, the combination word2vec-NB obtained the lowest results regarding overall metrics in the English dataset and the contrast between macro values and accuracy in the Spanish dataset. The BERT-NB combination also suffers the previous contrast. In order to use word embedding vectors, we changed the actual data distribution to the Gaussian distribution, which can disturb the distances between classes, causing more overlapping. This fact, along with an imbalanced dataset entail a high accuracy with classes with more emails, like Sexual Content Dating, and, in consequence, high general accuracy. However, the macro metrics show the poor results by considering all classes with the same weight.

**Table 3**
Performance of the sixteen pipelines in **P**recision, **R**ecall and **F1** Score using **Av**erage Macro, Micro and Weighted, **ACC**uracy and **R**un**T**ime (ms/email) terms.

| Pipeline/Metrics | | SPEMC-15K-E | | | | | SPEMC-15K-S | | | | |
|---|---|---|---|---|---|---|---|---|---|---|---|
| | | Avg Macro | Avg Micro | Avg Weighted | ACC (%) | RT | Avg Macro | Avg Micro | Avg Weighted | ACC (%) | RT |
| TF-IDF-SVM | P | 0.965 | 0.924 | 0.931 | | | 0.952 | 0.983 | 0.984 | | |
| | R | 0.923 | 0.924 | 0.924 | 92.4 | 98.6 | 0.940 | 0.983 | 0.983 | 98.3 | 59.8 |
| | F1 | 0.941 | 0.924 | 0.926 | | | 0.945 | 0.983 | 0.983 | | |
| TF-IDF-NB | P | 0.957 | 0.934 | 0.936 | | | 0.962 | 0.985 | 0.985 | | |
| | R | 0.858 | 0.934 | 0.934 | 93.4 | 3.1 | 0.933 | 0.985 | 0.985 | **98.5** | 3.4 |
| | F1 | 0.883 | 0.934 | 0.934 | | | **0.945** | **0.985** | **0.985** | | |
| TF-IDF-RF | P | 0.960 | 0.925 | 0.931 | | | 0.945 | 0.982 | 0.945 | | |
| | R | 0.906 | 0.925 | 0.925 | 92.1 | 40.5 | 0.941 | 0.982 | 0.941 | 98.2 | 8.7 |
| | F1 | 0.929 | 0.925 | 0.927 | | | 0.943 | 0.982 | 0.943 | | |
| TF-IDF-LR | P | 0.971 | 0.946 | 0.949 | | | 0.953 | 0.983 | 0.984 | | |
| | R | 0.939 | 0.946 | 0.946 | **94.6** | **2.0** | 0.944 | 0.983 | 0.983 | 98.3 | **2.2** |
| | F1 | **0.953** | **0.946** | **0.947** | | | 0.947 | 0.983 | 0.983 | | |
| BOW-SVM | P | 0.964 | 0.934 | 0.936 | | | 0.942 | 0.982 | 0.982 | | |
| | R | 0.923 | 0.934 | 0.934 | 93.4 | 55.8 | 0.937 | 0.982 | 0.982 | 98.2 | 48.0 |
| | F1 | 0.941 | 0.934 | 0.934 | | | 0.939 | 0.982 | 0.982 | | |
| BOW-NB | P | 0.675 | 0.890 | 0.918 | | | 0.809 | 0.932 | 0.962 | | |
| | R | 0.827 | 0.890 | 0.890 | 89.0 | 4.5 | 0.768 | 0.932 | 0.932 | 93.2 | 3.8 |
| | F1 | 0.682 | 0.890 | 0.900 | | | 0.734 | 0.932 | 0.930 | | |
| BOW-RF | P | 0.957 | 0.925 | 0.930 | | | 0.950 | 0.983 | 0.984 | | |
| | R | 0.908 | 0.925 | 0.925 | 92.5 | 35.6 | 0.946 | 0.983 | 0.983 | 98.3 | 9.0 |
| | F1 | 0.929 | 0.925 | 0.926 | | | 0.948 | 0.983 | 0.983 | | |
| BOW-LR | P | 0.966 | 0.943 | 0.945 | | | 0.947 | 0.983 | 0.984 | | |
| | R | 0.937 | 0.943 | 0.943 | 94.3 | 2.7 | 0.948 | 0.983 | 0.983 | 98.3 | 3.1 |
| | F1 | 0.950 | 0.943 | 0.944 | | | 0.947 | 0.983 | 0.983 | | |
| word2vec-SVM | P | 0.269 | 0.332 | 0.565 | | | 0.422 | 0.345 | 0.479 | | |
| | R | 0.402 | 0.332 | 0.315 | 33.2 | 77.5 | 0.557 | 0.345 | 0.345 | 34.5 | 49.2 |
| | F1 | 0.264 | 0.332 | 0.378 | | | 0.442 | 0.345 | 0.372 | | |
| word2vec-NB | P | 0.056 | 0.001 | 0.050 | | | 0.051 | 0.471 | 0.242 | | |
| | R | 0.125 | 0.001 | 0.012 | 1.4 | 3.7 | 0.097 | 0.471 | 0.471 | 47.1 | 5.0 |
| | F1 | 0.012 | 0.001 | 0.007 | | | 0.067 | 0.471 | 0.320 | | |
| word2vec-RF | P | 0.750 | 0.624 | 0.621 | | | 0.775 | 0.792 | 0.765 | | |
| | R | 0.346 | 0.624 | 0.623 | 62.4 | 17.0 | 0.654 | 0.792 | 0.792 | 79.2 | 25.2 |
| | F1 | 0.427 | 0.624 | 0.561 | | | 0.688 | 0.792 | 0.758 | | |
| word2vec-LR | P | 0.258 | 0.405 | 0.506 | | | 0.371 | 0.495 | 0.487 | | |
| | R | 0.397 | 0.405 | 0.405 | 40.5 | 12.2 | 0.462 | 0.495 | 0.495 | 49.5 | 8.9 |
| | F1 | 0.271 | 0.405 | 0.439 | | | 0.337 | 0.495 | 0.424 | | |
| BERT-SVM | P | 0.932 | 0.937 | 0.939 | | | 0.926 | 0.979 | 0.980 | | |
| | R | 0.951 | 0.937 | 0.937 | 93.7 | 8.7 | 0.924 | 0.979 | 0.979 | 97.9 | 4.0 |
| | F1 | 0.941 | 0.937 | 0.938 | | | 0.925 | 0.979 | 0.979 | | |
| BERT-NB | P | 0.413 | 0.614 | 0.717 | | | 0.352 | 0.823 | 0.833 | | |
| | R | 0.412 | 0.614 | 0.614 | 61.4 | 3.5 | 0.367 | 0.823 | 0.823 | 82.3 | 2.6 |
| | F1 | 0.358 | 0.614 | 0.621 | | | 0.297 | 0.823 | 0.811 | | |
| BERT-RF | P | 0.966 | 0.930 | 0.933 | | | 0.931 | 0.974 | 0.977 | | |
| | R | 0.898 | 0.930 | 0.931 | 93.0 | 39.4 | 0.895 | 0.974 | 0.974 | 97.4 | 7.8 |
| | F1 | 0.925 | 0.930 | 0.930 | | | 0.908 | 0.974 | 0.974 | | |
| BERT-LR | P | 0.932 | 0.942 | 0.943 | | | 0.897 | 0.979 | 0.979 | | |
| | R | 0.949 | 0.942 | 0.942 | 94.2 | 26.1 | 0.924 | 0.979 | 0.979 | 97.9 | 16.1 |
| | F1 | 0.939 | 0.942 | 0.943 | | | 0.908 | 0.979 | 0.979 | | |

Despite working with lower dimensionality vectors, the word embedding techniques do not overcome the processing time of term-frequency models for spam email classification. For instance, the BOW and TF-IDF combinations with LR are remarkably faster in comparison with word2vec and BERT pipelines. The combinations of SVM and LR with BERT, which is an encoder based on Deep Learning, are very close to term frequency encoders. However, they do not yield the best results and, due to their higher runtime –4ms in BERT-SVM and 16.1ms in BERT-LR against 2.2ms of TF-IDF-LR–, we do not select them for this application.

We present the confusion matrix for the highest-accuracy performance models for both languages in Figure 6. The classes with the worst accuracy are *Service* in the English dataset, confused with *Health* and *Other*, and *Identity Fraud* in the Spanish dataset, mislabelled with *Academic Media*, *Other* and *Work Offer*. Although, due to their thematic variety, the category *Other* presents a major confusion with other categories in both languages.

The per-class accuracy metric for the pipelines with the English dataset is shown in Figure 7. These graphics show that the classes *Health*, *Other* and *Services* are the ones with the lowest performance in the 16 pipelines. These three classes contain emails with a similar writing style, which may explain the problem description, relevance and the proposed solution. This email structure causes that the emails share similar words in their content, varying the thematic words. The category *Other* comprises a wide range of products and tricks, which results in not having a set of specific words to define it. This feature leads to this category to intersect with the rest of the classes. The category *Service* has a small number of examples, compared to *Health* and *Other*, so the pipelines based on term frequency might not differ among them.

There are classes with high accuracy in most of the pipelines: *Extortion Hacking*, *Pharmacy*, *Sexual Content Dating* and *Work Offer*. One of the reasons to explain this high performance in *Sexual Content Dating* is that it is the most representative class with 7924 emails. However, the remaining categories have a small number of emails. In consequence, the reason might be a robust set of representative words or repetitive email structure, for frequency-based and context-based text representation models, respectively. Finally, *Identity Fraud* is classified with high accuracy in most pipelines, being BOW the text representation technique with the best performance. Due to this, it might be emphasised the importance of word count to detect identity frauds in English emails.

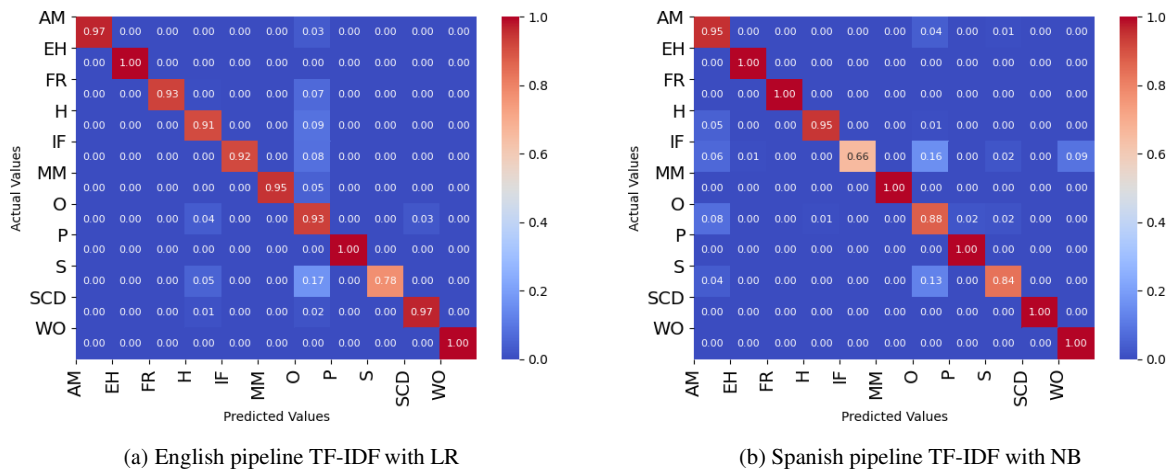

(a) English pipeline TF-IDF with LR  (b) Spanish pipeline TF-IDF with NB

**Fig. 6:** Confusion Matrix of the highest performance pipelines per language in terms of accuracy (%). We use the following acronyms Academic Media (AM), Extortion Hacking (EH), Fake Reward (FR), Health (H), Identity Fraud (IF), Money Making (MM), Other (O), Pharmacy (P), Services (S), Sexual Content Dating (SCD) and Work Offer (WO)

Regarding text representation models, the term frequency algorithms (TF-IDF and BOW) achieve similar performance in every class. It is worth highlighting the low accuracy of BOW-NB combination in *Academic Media*. The long extension of the emails implies more words alongside the NB principle of independence between features might produce the confusion with other classes with long emails. Also, the combination TF-IDF-NB obtained a low performance in Pharmacy class. The word2vec pipelines improve their performance in classes whose content is repetitive for most emails, such as Extortion Hacking, Money Making or Sexual Content Dating. BERT pipelines, except NB model, outperform the other ones in Academic Media, which contains long scientific emails with a formal expression and phrase constructions.

Figure 8 depicts the accuracy metric per class for Spanish pipelines. *Academic Media*, *Identity Fraud*, *Other* and *Services* are the classes with the worst overall results. This confusion might be explained by similar reasons to English. Particularly, the class *Other* also contains a wide range of topics that might not define it robustly. In Spanish datasets, *Identity Fraud* contains emails focused on company impersonation, what might interfere with the class *Service*. As well as it happens with *Academic Media* and *Service*, Spanish *Academic Media* emails involve many training courses from universities or academies that have similarities with *Service* emails.

On the other hand, the classes *Extortion Hacking*, *Money Making*, *Pharmacy*, *Sexual Content Dating* and *Work Offer* achieve generally a high performance. As it happens with the English pipelines, the number of examples, a well-defined representative set of words and similar email structure might be the reasons. The word2vec-based combinations, in English, obtain higher results in classes which contain emails with repetitive structure, such as *Money Making* or *Extortion Hacking*. The BOW pipelines also stand out to detect Identity Fraud class, what might indicate that the counting of words is important for Spanish multi-classification. Nevertheless, the performance in Spanish is lower than in English. This fact might have relation with a less number of examples and the kind of fraud different from English hoax. Moreover, the combination of BOW with NB decreases its performance in three classes, which are *Academic Media*, *Fake Reward*, and *Other*. Although the extension of the emails is short in these cases, the reasons might be the same as in English.

Since the class imbalance may affect negatively the performance of our models, we have carried out an analysis of alternatives to try to overcome this issue. Our datasets contain approximately 15K spam emails unevenly distributed in eleven classes and quite unbalanced. Besides the weighted class approach, we have evaluated two well-known combinations of over- and under-sampling: (i) Random over- and under-sampling, and (ii) Synthetic Minority Oversampling Technique (SMOTE) [13] along with Near-Miss [54] as undersampler. We performed the aforementioned over/under sampling strategies to balance the dataset resulting in a dataset with approximately the same size total size of 15k spam emails. We show the comparison to our previous results using class weight approach and both over-/under-sampling strategies in Table 4. We can observe that applying a class weighted method or sampling strategy (guided by SMOTE and NearMiss) outperforms the solution achieved with a random sampling strategy in order to balance the dataset.

**Table 4**
Performance of the three imbalanced alternatives applied to the hightes performance models, TF-IDF-LR (English) and TF-IDF-NB (Spanish). in **P**recision, **R**ecall and **F1** Score using **Av**erage Macro, Micro and Weighted, and **ACC**uracy terms.

| | | TF-IDF-LR trained in SPEMC-15K-E | | | | TF-IDF-NB trained in SPEMC-15K-S | | | |
|---|---|---|---|---|---|---|---|---|---|
| Pipeline/ Metrics | | Avg Macro | Avg Micro | Avg Weighted | ACC (%) | Avg Macro | Avg Micro | Avg Weighted | ACC (%) |
| Class weight | P | 0.971 | 0.946 | 0.949 | | 0.962 | 0.985 | 0.985 | |
| | R | 0.939 | 0.946 | 0.946 | 94.6 | 0.933 | 0.985 | 0.985 | 98.5 |
| | F1 | 0.953 | 0.946 | 0.947 | | 0.945 | 0.985 | 0.985 | |
| Over-Under-sampling | P | 0.922 | 0.914 | 0.922 | | 0.978 | 0.978 | 0.978 | |
| | R | 0.914 | 0.914 | 0.914 | 91,4 | 0.978 | 0.978 | 0.978 | 97,8 |
| | F1 | 0.914 | 0.914 | 0.914 | | 0.978 | 0.978 | 0.978 | |
| SMOTE+NearMiss | P | 0.971 | 0.967 | 0.971 | | 0.968 | 0.967 | 0.968 | |
| | R | 0.967 | 0.967 | 0.967 | 96,7 | 0.967 | 0.967 | 0.967 | 96,7 |
| | F1 | 0.967 | 0.967 | 0.967 | | 0.967 | 0.967 | 0.967 | |

Additionally, we compared our proposed method with the class detection technique used in [44], that matches the most frequent words in a category. The spam keywords used are the 18 most frequent words presented in Section 3. We obtained a 66.7% and 95.9% of accuracy for SPEMC-15-E and SPEMC-15-S, respectively. Although this method provides quite remarkable performance in the Spanish dataset, the poor results in the English dataset show that it has a high dependency on the similarity among emails from the same dataset. Our proposal outperforms in terms of accuracy both language scenarios with 94.6% in SPEMC-15K-E and 98.5% in SPEMC-15K-S.

## 6. Conclusions

In this work, we addressed the problem of spam email analysis following a multi-class classification approach, focusing on the needs of a cybersecurity institute. With the aim to extract relevant information from massive amounts of spam emails, we categorise spam emails using NLP and machine learning techniques. To the best of our knowledge, our work is among the first to address the problem of spam topics with the purpose of carrying out an advanced analysis of its content.

We created SPEMC-15K-E and SPEMC-15K-S, two novel datasets which contain 14479 English and 14992 Spanish spam emails, respectively. We semi-automatically labelled them into eleven spam categories according to their topic, using hierarchical clustering first and, later, manual inspection supported by cybersecurity experts.

Additionally, we detected the spammer trick known as hidden text or "salting" inside some emails. We solved it by converting the HTML email into an image and then extracting with an OCR the textual content that is visible for the user. We also addressed the problem of the spam contained in images attached to an email by extracting the text inserted into the images using an OCR. This approach mitigates the confusion made by both spam filters and spam multi-classifiers.

In order to categorise spam into eleven classes, we assessed the combination of four text representation techniques (frequency-based and word embedding-based) with four traditional machine learning algorithms, resulting in 16 pipelines and recommending the best combination for each language. We evaluated each pipeline in terms of macro precision, recall, and F1 score as well as accuracy and run time.

Most pipelines achieved high overall performance, but the frequency-based text representation models, TF-IDF and BOW, generally outperformed the word embedding models in the spam multi-classification task, being also lighter and quicker models.

Considering the metrics F1 score and accuracy, the combination of TF-IDF with LR obtained the highest performance with 0.953 of F1 score and 94.6% of accuracy on SPEMC-15K-E and 0.945 and 98.5% on SPEMC-15K-S, respectively. Regarding the run-time per email, we also recommend the combination of TF-IDF with LR for real-time application on English and Spanish spam multi-classification. They classify an English email into one of 11 classes in 2 ms and a Spanish email in 2.2 ms, on average.

For the next stage of our research, we are interested in looking for more relevant features, alternatives to use NB with negative values and testing pre-trained models for word2vec in order to improve their performance. Moreover, testing other lighter models such as ALBERT or ELECTRA, which may enhance both accuracy and runtime, becomes part of our immediate future research. Experimental results encourage us to deepen in the characteristics of each class in order to detect patterns that help identify campaigns against companies, citizens privacy and security. We will also seek to find associations between classes and spam tricks, which help identify organisations behind campaigns.

## 7. Acknowledgments

This work was supported by the framework agreement between the Universidad de León and INCIBE (Spanish National Cybersecurity Institute) under Addendum 01.

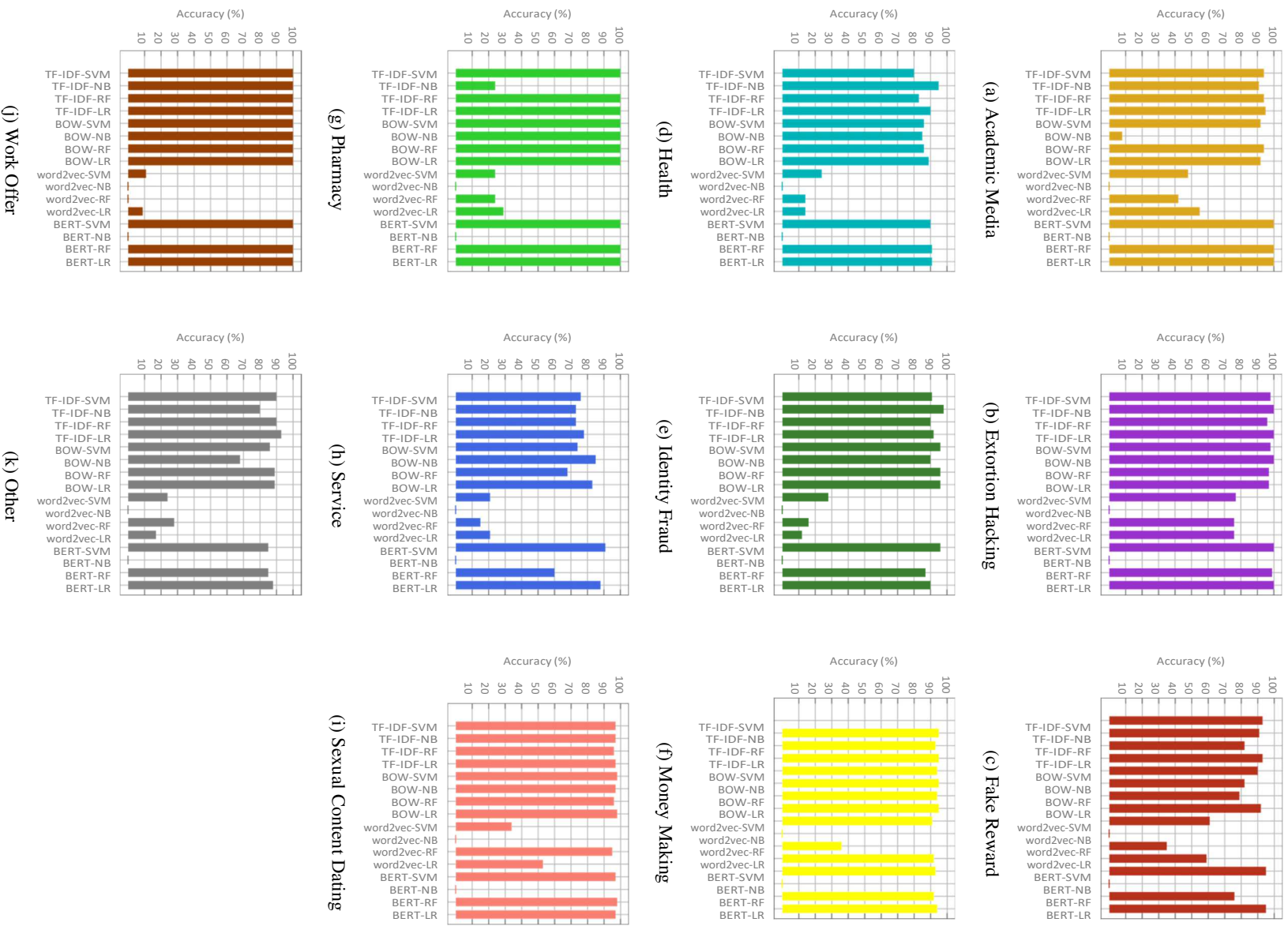

**Fig. 7**: Performance of every English pipeline per category in term of accuracy (%).

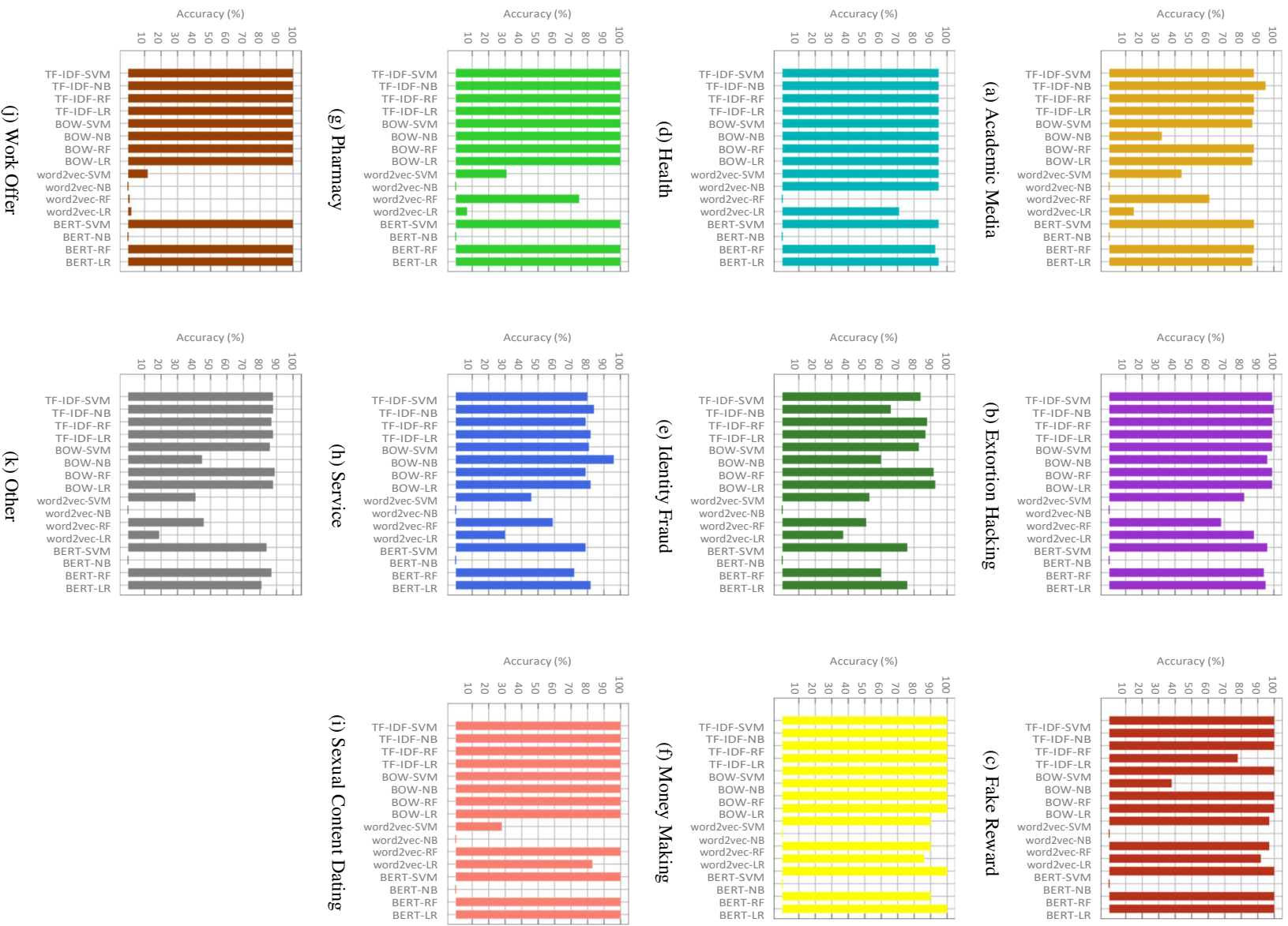

**Fig. 8**: Performance of every Spanish pipeline per category in terms of accuracy (%).